%% file: main.tex
\documentclass[10pt, conference]{IEEEtran}
\IEEEoverridecommandlockouts
\usepackage{cite}
\usepackage{amsmath,amssymb,amsfonts, amsthm}
\usepackage{algorithmic}
\usepackage{algorithm}
\usepackage{graphicx}
\usepackage{svg}
\usepackage{textcomp}
\usepackage{xcolor}
\usepackage{mathtools}
\usepackage{booktabs}
\usepackage{multirow}
\usepackage{longtable}
\usepackage{listings}
\usepackage{soul} %
\usepackage[colorlinks]{hyperref}
\usepackage[acronym,nonumberlist]{glossaries}

\makeglossaries
\usepackage{framed,enumitem}

\hypersetup{
    colorlinks  = true,
    linkcolor   = black,     %
    anchorcolor = black,
    citecolor   = green,   %
    filecolor   = cyan,
    menucolor   = red,
    runcolor    = cyan,
    urlcolor    = blue    %
}
\def\BibTeX{{\rm B\kern-.05em{\sc i\kern-.025em b}\kern-.08em
    T\kern-.1667em\lower.7ex\hbox{E}\kern-.125emX}}
\newcommand{\etal}{\textit{et al.}}
\renewcommand{\S}{Section }

\theoremstyle{definition}
\newtheorem{definition}{Definition} 

\newacronym{nserc}{NSERC}{Natural Sciences and Engineering Research Council of Canada}
\newacronym{oce}{OCE}{Ontario Center of Excellence}
\newacronym{dl}{DL}{Deep Learning}
\newacronym{od}{OD}{Object Detection}
\newacronym{cnn}{CNN}{Convolutional Neural Network}
\newacronym{yolo}{YOLO}{You Only Look Once}
\newacronym{covid}{COVID-19}{Coronavirus Disease}
\newacronym{gpu}{GPU}{Graphical Processing Unit}
\newacronym{rcnn}{R-CNN}{Region Proposal Convolutional Neural Network}
\newacronym{fpn}{FPN}{Feature Pyramid Network}
\newacronym{cspnet}{CSPNet}{Cross-Stage Partial Connections}
\newacronym{panet}{PANet}{Path Aggregation Network}
\newacronym{sam}{SAM}{Spatial Attention Module}
\newacronym{ciou}{CIoU}{Complete Intersection Over Union}
\newacronym{silu}{SiLU}{Sigmoid-weighted Linear Unit}
\newacronym{elan}{ELAN}{Efficient Long-Range Attention Network}
\newacronym{repconvn}{RepConvN}{Reparameterized Convolutional}
\newacronym{dfl}{DFL}{Distribution Focal Loss}
\newacronym{lfyolo}{LF-YOLO}{Lighter and Faster YOLO}
\newacronym{rmf}{RMF}{Reinforced Multiscale Feature}
\newacronym{attyolo}{ATT-YOLO}{Attention YOLO}
\newacronym{senet}{SENet}{Squeeze and Excitation Network}
\newacronym{yoloimf}{YOLO-IMF}{YOLO for Industrial Manufacturing Field}
\newacronym{mscoco}{MS-COCO}{Microsoft Common Objects in Context}
\newacronym{rtx}{RTX}{Ray Tracing Texel eXtreme}
\newacronym{p}{P}{Precision}
\newacronym{r}{R}{Recall}
\newacronym{tp}{TP}{True Positive}
\newacronym{fp}{FP}{False Positive}
\newacronym{tn}{TN}{True Negative}
\newacronym{fn}{FN}{False Negative}
\newacronym{map}{mAP}{Mean Average Precision}
\newacronym{iou}{IoU}{Intersection over Union}
\newacronym{flops}{FLOPS}{Floating-Point Operations Per Second}
\newacronym{id}{ID}{Identifier}
\newacronym{deepsort}{DeepSort}{Deep Simple Object Tracking}
\newacronym{kf}{KF}{Kalman Filtering}
\newacronym{mes}{MES}{Manufacturing Execution System}
\newacronym{lmm}{LMM}{Large Multimodal Model}

\begin{document}
\title{Capacity Constraint Analysis Using Object Detection for Smart Manufacturing
\thanks{\IEEEauthorrefmark{1} Corresponding author: ahmad54@uwindsor.ca \\
This study is supported by IFIVEO CANADA INC., Mitacs through IT16094, \acrfull{nserc} through ALLRP 560406-20, \acrfull{oce} through OCI\# 34166, and the University of Windsor, Canada.
}
}
\author{
    \IEEEauthorblockN{
        Hafiz Mughees Ahmad\IEEEauthorrefmark{1}\IEEEauthorrefmark{2},
        Afshin Rahimi\IEEEauthorrefmark{2},
        Khizer Hayat\IEEEauthorrefmark{3}
    \IEEEauthorblockA{
        \IEEEauthorrefmark{2}\textit{Mechanical, Automotive and Materials Engineering Department, University of Windsor, Windsor, ON, Canada} \\
        \{ahmad54, arahimi\}@uwindsor.ca}
    \IEEEauthorblockA{
        \IEEEauthorrefmark{3}\textit{IFIVEO CANADA INC. Windsor, ON, Canada} \\
        khizer@ifiveo.com}
    }
}
\maketitle
\begin{abstract}
The increasing popularity of \acrfull{dl} based \acrfull{od} methods and their real-world applications have opened new venues in smart manufacturing. Traditional industries struck by capacity constraints after \acrfull{covid} require non-invasive methods for in-depth operations' analysis to optimize and increase their revenue. In this study, we have initially developed a \acrfull{cnn} based \acrshort{od} model to tackle this issue. This model is trained to accurately identify the presence of chairs and individuals on the production floor. The identified objects are then passed to the \acrshort{cnn} based tracker, which tracks them throughout their life cycle in the workstation. The extracted meta-data is further processed through a novel framework for the capacity constraint analysis. We identified that the Station C is only 70.6\% productive through 6 months. Additionally, the time spent at each station is recorded and aggregated for each object. This data proves helpful in conducting annual audits and effectively managing labor and material over time.
\end{abstract}


\section{Introduction}
The manufacturing sector has long been a cornerstone of economic development, driving innovation and providing employment opportunities. However, the outbreak of the \acrshort{covid} pandemic has exacerbated existing challenges in the manufacturing landscape, notably the critical issue of labor shortage and supply chain collectively referred to as capacity constraint. Industries have difficulty finding skilled labor for tasks requiring precise human effort. According to Causa \etal \cite{causa_covid_2022}, 75\% of employers have a hard time filling open positions, and manufacturing is among the most hit. Additionally, as per the \textit{Analysis on labor challenges in Canada, second quarter of 2023} by \textit{Statistics Canada} in June 2023 \cite{government_of_canada_analysis_2023}, 59.3\% of manufacturing industries consider rising inflation to be an obstacle over the next three months. Nearly 9 out of 10 organizations surveyed responded that they are having a hard time filling open positions, which are mostly comprised of entry-level to mid-level positions, with manufacturing being the most hit as 93\% struggled to find entry-level employees. The Canadian Federation of Independent Business also reported \cite{bomal_laure-anna_labour_nodate} that as of November 2021, 55\% of small businesses in Canada experienced labor shortage and difficulty in hiring, retaining or getting staff to work the needed hours. Overall, this shortage is exponentially increasing the existing global supply chain issues. 

This confluence of data from various sources underscores a consistent and pressing issue of capacity constraint that requires innovative solutions. Researchers and engineers trying to solve this unique challenge have started consulting to use innovative methods to increase labor productivity and decrease the bottlenecks in the production pipeline to reduce the effect of the supply chain \cite{gervasi_applications_2023, poudel_decentralized_2023, liu_effects_2023, pansara_fields_2023}. Based on an earlier study by Ahmad and Rahimi \cite{ahmad_deep_2022}, \acrshort{od} applications in smart manufacturing play a pivotal role in enhancing quality control, cycle-time studies, safety compliance, and surveillance. Puttemans \etal, \cite{puttemans_building_2020} and Wang \etal \cite{wang_machine_2019} employed deep learning based \acrshort{od} model \cite{redmon_yolo9000_2017} for detecting packages in warehouse environments, highlighting its utility in real-time applications for product packaging. Farahnakian \etal \cite{farahnakian_towards_2021} along with Li \etal \cite{li_application_2019} applied \acrshort{od} models for damage detection and pallet rack identification in industrial warehouse settings.

While much of the existing research has concentrated on automating manufacturing processes, the field of manual production, particularly where the productivity of skilled labor is paramount, remains relatively unexplored. In response to the complex challenges faced by various sectors within the manufacturing industry, this paper introduces a collaborative research initiative involving a computer vision company, IFIVEO CANADA INC., and its client (hereafter referred to as \textit{the client} in this article), which specializes in the production of assistive medical wheelchairs. The manual assembly of power-assist wheelchairs, a niche yet crucial manufacturing segment, presents unique challenges. These wheelchairs, custom-built based on medical prescriptions, require meticulous precision and expertise. 

This article delves into the innovative use of \acrshort{cnn}s to identify bottlenecks in the production pipeline, a pervasive issue worsened by labor shortages and faulty parts. Our approach begins with developing and training a \acrshort{cnn} based \acrshort{od} model meticulously designed to identify power-assist wheelchairs, referred to as \textit{chairs} henceforth, and workers on the production floor. Subsequently, a \acrshort{cnn} based tracking system is employed to keep track of their lifecycle. The extracted meta-data is then processed to provide insights about productivity in their manufacturing facility. The overall approach provides real-time metrics of the production capacity and can be further utilized for capacity constraint analysis. 

Our key contribution within this study can be summarized as a non-invasive state-of-the-art framework for analyzing the capacity of the manufacturing facility using \acrshort{od} methods which categorize the station into different states. This can be specially used in industries struck by capacity constraints and help them optimize the processes by removing the bottlenecks and increasing overall production and revenues in general.

The following sections delve into the state-of-the-art \acrshort{od} methods and their applications in smart manufacturing in \S \ref{sec:releted_work}. We propose a theoretical framework in \S \ref{sec:labour_productivity} while the technical intricacies of our proposed solution are supported by comprehensive analysis and empirical evidence in \S \ref{sec:methodology}. Insights from the manufacturing facility are discussed in \S \ref{sec:insights} and finally \S \ref{sec:conslusion} offers concluding remarks. 

\section{Related Work} \label{sec:releted_work}
\acrshort{cnn} based \acrshort{od} methods and their applications in real-world value-added services are an active area of research as these models are pivotal in the localization and classification of the objects within a given frame. 

Historically, traditional approaches relied on carefully engineered hand-crafted features, leading to time-consuming and less accurate results. However, the advent of \acrshort{cnn} based deep learning models, empowered by increased computational capacity through a \acrfull{gpu}, has revolutionized computer vision.

Two primary categories of \acrshort{od} methods have emerged: region proposal and regression based methods. \acrfull{rcnn}s \cite{girshick_rich_2014} propose regions that are subsequently classified into predefined categories \cite{girshick_fast_2015, ren_faster_2017}. While these models demonstrated high localization accuracy, they were computationally complex, often falling short of real-time performance due to the proposal of thousands of regions per image. To address this limitation, one-stage detectors were introduced \cite{liu_ssd_2016} including the groundbreaking \acrfull{yolo} by Redmon \etal \cite{redmon_you_2016}, which provided real-time performance across various benchmarks.

\subsection{You Only Look Once}
\acrshort{yolo} \cite{redmon_you_2016} represents a revolutionary approach to \acrshort{od} and localization, treating the problem as a regression task. In essence, \acrshort{yolo} directly proposes bounding box coordinates and associated class probabilities from the image pixels, presenting a unified and end-to-end trainable model. This monolithic architecture learns directly from the input images during training, eliminating the need for complex multi-stage pipelines.

The history of \acrshort{yolo} models in \acrshort{od} is marked by significant technical advancements \cite{terven_comprehensive_2023}. \acrshort{yolo}v1 \cite{redmon_you_2016} (2016) introduced a single end-to-end architecture that simultaneously predicted multiple bounding boxes and class probabilities for those boxes, which significantly improved the speed compared to previous region proposal based methods \cite{girshick_rich_2014}. The authors achieved this by dividing the input image into a grid, with each grid cell responsible for detecting objects within it. Each cell can predict multiple bounding boxes and confidence scores for those boxes. The network used a combination of 24 convolutional layers and 2 fully connected layers, with a final output tensor providing class probabilities and bounding box coordinates. 
\acrshort{yolo}v2 \cite{redmon_yolo9000_2017} (2017), and \acrshort{yolo}v3 \cite{redmon_yolov3_2018} (2018) brought improvements in both speed and accuracy by introducing the anchor boxes to predict offsets rather than the full bounding box and refining the feature extractor resulting in improved accuracy. The authors also propose Darknet-19 and Darknet-53 consisting of 19 and 53 layered networks, respectively, while incorporating multi-scale predictions using a \acrfull{fpn} \cite{lin_feature_2017} for improved detection of small objects.

\acrshort{yolo}v4 \cite{bochkovskiy_yolov4_2020} (2020) proposed the integration of \acrfull{cspnet} \cite{wang_cspnet_2020}, \acrfull{panet} \cite{liu_path_2018} and modified \acrfull{sam} \cite{woo_cbam_2018} along with the use of the Mish \cite{misra_mish_2020} activation function and \acrfull{ciou} loss \cite{zheng_distance-iou_2020} for enhancing feature extraction and bounding box accuracy.
\acrshort{yolo}v5 \cite{jocher_ultralyticsyolov5_2020} (2020) introduced a more streamlined and simplified architecture along with model scalability enhancements to adjust the model size based on the available computational resources. The authors improved the performance using novel \textit{Mosaic} augmentation combined with multiple other data augmentation methods for data pre-processing. This increased the variance in the data, hence improving the detection accuracy. They also used the \acrfull{silu} activation function \cite{elfwing_sigmoid-weighted_2017} instead of Mish \cite{misra_mish_2020} as employed by their predecessors. 
\acrshort{yolo}v6 \cite{li_yolov6_2022} (2021) and \acrshort{yolo}v7 \cite{wang_yolov7_2023} (2022) focused on optimizing the balance between speed and accuracy for edge computing devices. They used the \acrfull{elan} \cite{wang_designing_2022} strategy to increase its convergence speed and reduce the training time. The authors also proposed a \acrfull{repconvn} block inspired from \cite{ding_repvgg_2021} that helped better feature extraction.

\begin{figure*}[t]
    \centering
    \includegraphics[width=1\textwidth]{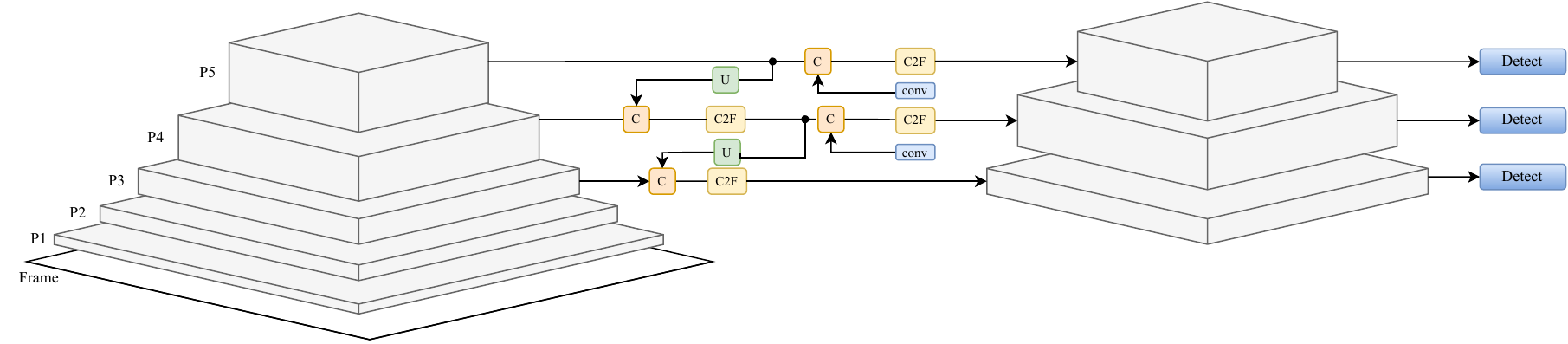}
    \caption{Complete \acrshort{yolo}v8 \cite{jocher_yolo_2023} architecture consisted of backbone and head. C represents the convolutional block, U is the upsampling block, and C2F is the \acrshort{cspnet} with two convolutional layers. A detailed diagram can be found in \cite{king_brief_2023}.}
    \label{fig:yolov8 architecture}
\end{figure*}

\acrshort{yolo}v8 \cite{jocher_yolo_2023} (2023), the latest in the series, represents the culmination of ongoing efforts to optimize \acrshort{od} for both performance and computational efficiency and advanced network architecture incorporating recent developments in neural network design. The authors used an anchor-free model inspired by \cite{ge_yolox_2021} with a decoupled head to independently process objectness, classification, and regression tasks along with \acrshort{ciou} \cite{zheng_distance-iou_2020} and \acrfull{dfl} \cite{li_generalized_2020} functions for bounding box loss and binary cross-entropy \cite{good_rational_1952} for classification loss. Figure \ref{fig:yolov8 architecture} illustrates the fully visualized \acrshort{yolo}v8 architecture, with different stages of the network shaded in distinct colors for clarity. 
The \acrshort{yolo} series continues to be a prominent example of innovation in computer vision and deep learning. Each version has contributed to the rapid progression and adaptability of deep learning models to become more efficient and capable, making them suitable for various applications, from mobile and edge devices to high-end \acrshort{gpu}s. 

\subsection{Object Detection in Smart Manufacturing}
The researchers have been actively identifying innovative ways to utilize the power of \acrshort{od} methods in smart manufacturing. Recently, Zendehdel \etal \cite{zendehdel_real-time_2023} used the \acrshort{yolo}v5 \cite{jocher_ultralyticsyolov5_2020} model to identify and localize the tools on the manufacturing floor for worker safety. Liu \etal \cite{liu_lf-yolo_2023} proposed a novel \acrfull{lfyolo} model for defect detection on the X-ray imagery of the welding where they also proposed \acrfull{rmf} module to extract more hierarchical information. Wang \etal \cite{wang_toward_2023} proposed a lightweight \acrshort{yolo} style object detector known as \acrfull{attyolo} for surface defect detection in electronics manufacturing. Zhao \etal \cite{zhao_rdd-yolo_2023} also modified \acrshort{yolo}v5 \cite{jocher_ultralyticsyolov5_2020} for the steel surface defects where they modified the architecture to utilize low-level features for better detection. Puttemans \etal \cite{puttemans_building_2020} and Vu \etal \cite{vu_yolo-based_2023} employed \acrshort{yolo}v2 \cite{redmon_yolo9000_2017} and \acrshort{yolo}v5 \cite{jocher_ultralyticsyolov5_2020} for detecting packages in warehouse environments, highlighting its utility in real-time applications for product packaging. Zhao \etal \cite{zhao_real-time_2021} proposed modified \acrshort{yolo}v5 for lightweight, real-time performance in detecting particleboard surface defects. This was achieved by replacing conventional convolutional layers with depth-wise convolution layers and integrating \acrfull{senet} \cite{hu_squeeze-and-excitation_2018} layers to optimize the model's parameters. Rahimi \etal \cite{rahimi_object_2021} proposed modifications to \acrshort{yolo}v3 for detecting large-scale objects, specifically in the automobile industry, enhancing the model by altering its architecture and activation function. Ahmad \etal \cite{ahmad_deep_2021} applied \acrshort{yolo}v3 for detecting and tracking cranes in steel manufacturing plants, showcasing the adaptability of these deep learning models for specific industrial surveillance tasks. Liu \etal \cite{liu_yolo-imf_2023} proposed \acrfull{yoloimf}, an improved \acrshort{yolo}v8 algorithm for surface defect detection in the industrial manufacturing field. Luo \etal \cite{luo_hardware-friendly_2023} modified \acrshort{yolo}v8 towards edge computing by reducing the parameters and computational load by modifying the lightweight ShuffleNetV2 network \cite{ma_shufflenet_2018} and using that as the feature extractor of \acrshort{yolo}. 

Additionally, Krummenacher \etal \cite{krummenacher_wheel_2018} applied deep learning for wheel defect detection. O’Brien \etal \cite{obrien_object_2019} introduced a method for inspecting the quality of medical device production, addressing the need for high accuracy and low error tolerance in applications for medical equipment manufacturing. Farahnakian \etal \cite{farahnakian_towards_2021} and Li \etal \cite{li_application_2019} used \acrshort{od} methods for damage detection and pallet rack identification in industrial warehouse settings. Wei \etal \cite{wei_deep_2018} and Luo \etal \cite{luo_benchmark_2019} demonstrated the effectiveness of deep learning in detecting humans and industrial tools from a distance, showcasing the versatility of these models in diverse industrial scenarios. Wang \etal \cite{wang_machine_2019} advanced product defect detection using a deep learning approach that synergies pre-processing techniques with deep learning, reducing computational load and excluding irrelevant background content.

With all these applications, it is evident that \acrshort{od} methods have been very effective and are gaining more popularity in the research and engineering community. Nevertheless, these are more focused on materials quality while the productivity of the labor handling such materials and workload is not studied and needs further and more thorough exploration. In the next section, we have defined an initial framework for this missing piece in the literature.  
\section{Capacity Constraint Analysis} \label{sec:labour_productivity}
Capacity constraint in a manufacturing environment is defined as a situation where a business's production capacity is insufficient to meet demand. This limitation can manifest in various forms, such as labor, materials, or equipment constraints. According to the Theory of Constraints \cite{goldratt_goal_2016, goldratt_theory_1990}, every system has bottlenecks that dictate the pace of the entire production line, and addressing these bottlenecks can significantly increase overall output. The initial step in this process involves thoroughly analyzing the constraints to pinpoint the actual limiting factors. In this section, we propose a comprehensive framework to analyze workstations, particularly those operated by manual labor. This framework aims to identify the root causes of capacity limitations and develop strategic, long-term solutions to enhance manufacturing efficiency and output.

To effectively analyze workstations and identify bottlenecks in the manufacturing process, it is essential to consider productivity, which is fundamentally defined as the output per unit of input. Productivity is a critical metric for operational efficiency, especially in contexts with limited labor resources. While organizations may use various parameters for this analysis, our proposed framework incorporates a holistic set of definitions considering both workforce and material. This integrated approach ensures a more accurate identification of constraints and facilitates the development of targeted solutions to optimize manufacturing processes.

\noindent\rule{\linewidth}{1pt}

\begin{definition}[Station Productivity]
We define station productivity as the output generated when a worker is actively engaged in working with materials. 
\end{definition}
This measure focuses on the station's effectiveness in producing items rather than solely assessing the worker's efficiency in the quantity of objects produced. It's a nuanced measure of efficiency, highlighting how effectively a station utilizes its resources (both human and material) to generate output.
\begin{definition}[Non-Productivity]
Non-productivity at a station occurs when no value is produced due to worker unavailability. 
\end{definition}
This typically happens when a worker is absent from their station. However, it's important to differentiate between avoidable and unavoidable non-productive time. Breaks, for example, are a necessary aspect of work that, while non-productive, contribute to overall worker productivity and well-being by preventing fatigue and maintaining mental health.
\begin{definition}[Downtime]
Downtime in this context refers to periods when a worker is present but lacks the necessary materials to continue production.
\end{definition}
This situation can arise due to supply chain issues, scheduling errors, or unforeseen delays in material delivery. Downtime is a critical aspect of station productivity as it directly impacts the output despite the availability of workers.
\begin{definition}[Idle Time]
Idle time is characterized by the absence of both workers and materials at a station. 
\end{definition}
This occurs during off-hours or designated break times. Understanding idle time is crucial for workforce planning and ensuring that staffing levels are appropriate to the demands of the production schedule.

\noindent\rule{\linewidth}{1pt}

Mathematically, all of the above definitions can be combined and represented as a simple lookup table shown in Table \ref{tab:productivity_chart} to get the status of each frame.
\begin{table}[htb]
\centering
\caption{Different possibilities for the station productivity. The \checkmark represents the availability at the station. }
\label{tab:productivity_chart}
\begin{tabular}{lcc}
\toprule 
                & Material   & Worker     \\ \midrule 
Productive      & \checkmark & \checkmark \\ \hline 
Un-productivity & \checkmark &            \\ \hline
Downtime        &            & \checkmark \\ \hline
Idle-time       &            &            \\ \bottomrule
\end{tabular}
\end{table}
\subsection{Cycle-time Study}
We can get the station status for each frame; however, aggregating that status over time provides the real value for the life of an object in the scene, also known as cycle time study. We need to track when an object appears in and leaves the scene. This can be measured using several methods. Here, we discuss several of these methods with their advantages and disadvantages as follows, 
\begin{itemize}[align = left, wide=0pt, leftmargin=2em]
    \item[\textbf{Stop Watches}] involves timing tasks manually. It's common but limited by sample size and may not accurately represent normal working conditions.
    \item[\textbf{Video Recording with Offline Analysis}] allows for efficient analysis but suffers from delays in feedback.
    \item[\textbf{Breaking Up Activity into Tasks and Subtasks}] helps in understanding task performance and supports line balancing, but it can be time-consuming.
    \item[\textbf{Working with Predetermined Standard Times}] offers deep insights but may not align with real-world timings.
    \item[\textbf{Sensor based Tracking}] involves using data from workflow systems for real-time productivity analysis. While efficient, it lacks insights into the causes of productivity changes.
    \item[\textbf{Visual Tracking}] combines real-time, non-intrusive data collection with identifying improvement opportunities. It requires upfront investment but is increasingly cost-effective.
\end{itemize}
While most methods require manual calculation and human input, visual tracking is real-time and does not need human feedback once the development is complete and will yield accurate results. It can be implemented by \acrshort{od} based methods and can be used to accurately conduct this cycle time effectively. Within this proposed methodology, we will use the \acrshort{yolo}v8 model as the state-of-the-art \acrshort{od} model. 
\section{Methodology} \label{sec:methodology}
The proposed methodology consists of a two-step approach. 1) We need data for training an \acrshort{od} model, and in the preceding section, we highlighted the state-of-the-art performance of \acrshort{yolo}v8 and decided to use that as our primary \acrshort{od} model in this study. 2) Testing that model in the manufacturing facility of choice to conduct the capacity constraint analysis. 
\subsection{Dataset Description}
For the dataset employed in the training and evaluation of the \acrshort{od} model, manual annotation was carried out on videos sourced directly from the four production line stations within the clients' facility over roughly two months. The facility runs only for one shift of 8.5 hours every day. A total of 33,956 individual frames were extracted from these recordings at a frequency of 0.3 frames per second ($fps$). As the scene in each station does not change, only the objects move; we needed manual stratified data splitting into train and validation to avoid data leakage as data from each day should only be in either split. The total images in train/validation ended up as 29,070/4,886. The dataset consisted of 2 classes, i.e., worker and chair. Each station has a different viewing angle, adding variation to the dataset.
\subsection{Training}
To train the \acrshort{yolo}v8 model as defined by \cite{jocher_yolo_2023}, we employed a transfer learning approach to refine the pre-trained model, initially trained on the \acrfull{mscoco} dataset \cite{lin_microsoft_2015}. The open-source model implementation\footnote{we have used the open source implementation of \acrshort{yolo}v8 model available at \url{https://github.com/ultralytics/ultralytics}} is available in 5 different sizes; nano (n), small (s), medium (m), large (l) and extra-large (x) depending on the number of parameters. Table \ref{tab:yolov8_metrics} represents the number of parameters in each model. We trained nano, medium, and large as three separate models to compare accuracy and detection speed. The model's output layer was reconfigured to identify two distinct object types. To enhance accuracy, non-maximum suppression \cite{neubeck_efficient_2006} was utilized for output refinement. Moreover, the cosine annealing learning rate method \cite{loshchilov_sgdr_2016} was implemented as a scheduler chosen for its demonstrated excellence in various benchmark tests.

To augment our dataset, we incorporated several methods, including random vertical flipping and a mosaic of 4 frames. These techniques expanded the dataset size and contributed to a more robust training process. The training was done using 2x NVIDIA \acrfull{rtx} TITAN \acrshort{gpu}s with a batch size of 128 for nano and medium and 64 for large-size models due to \acrshort{gpu} memory limitation. 
\begin{figure}
    \centering
    \includegraphics[width=0.5\linewidth,angle=0]{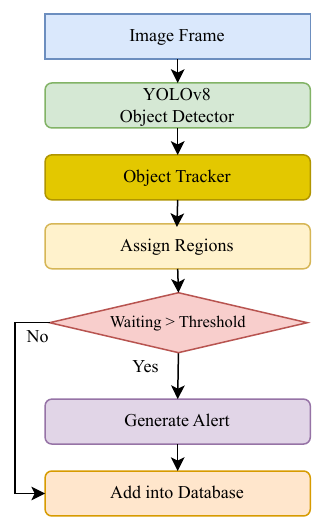}
    \caption{Complete pipeline of the proposed methodology. }
    \label{fig:pipeline}
\end{figure}
\subsection{Evaluation Metrics}
Detection accuracy and inference speed are key metrics for evaluating the \acrshort{od} model. Accuracy is often evaluated using \acrfull{p}  and \acrfull{r}, which are derived from the counts of \acrfull{tp}, \acrfull{fp}, \acrfull{tn}, and \acrfull{fn} while speed is measured in $fps$. The formulas for these metrics are as follows:
\begin{gather} %
\text{R} = \frac{\text{TP}}{\text{TP} + \text{FN}} \\
\text{P} = \frac{\text{TP}}{\text{TP} + \text{FP}}
\end{gather} 
Furthermore, \acrfull{map} assesses detection accuracy. This measure is calculated for each class based on \acrshort{p} and \acrshort{r} and then averaged to yield an overall score. For quantifying the accuracy of object localization, the \acrfull{iou} metric is used and calculated between the labeled objects (ground truth) and the model's predictions as follows:
\begin{equation}
\text{IoU} = \frac{\text{Area}(b_{pred} \cap b_{g})}{\text{Area}(b_{pred} \cup b_{g})}
\end{equation}
where \(b_g\) represents the ground truth bounding box, and \(b_{pred}\) denotes the bounding box predicted by the \acrshort{od} model. The \acrshort{iou} threshold functions as a boolean operator to eliminate \acrshort{fp} bounding boxes that score below a certain \acrshort{iou} value. This threshold determines the necessary sensitivity for the localization to be classified as positive or negative (e.g., \acrshort{iou} \(\geq\) threshold). Different models may employ varying threshold values, such as 0.25, 0.5, or 0.75, in their evaluations. Table \ref{tab:yolov8_metrics} lists the evaluation results and inference speed of the different \acrshort{yolo}v8 models where operations are quantified by \acrfull{flops}. 
\begin{table}[!tb]
\centering
\caption{Different model sizes of \acrshort{yolo}v8 Model from \cite{jocher_yolo_2023}. \acrshort{map} values are calculated for single-scale on \acrshort{mscoco} \cite{lin_microsoft_2015} val2017 dataset.}
\label{tab:yolov8_metrics}
\begin{tabular}{lcccccc}
\toprule Model & \acrshort{map}$_{0.50-0.95}$ & \acrshort{gpu} (ms) & Parameters (M)\\
\midrule
\acrshort{yolo}v8-n & 18.4 & 1.21 &    3.5 \\
\acrshort{yolo}v8-s & 27.7 & 1.40 &   11.4 \\
\acrshort{yolo}v8-m & 33.6 & 2.26 &   26.2 \\
\acrshort{yolo}v8-l & 34.9 & 2.43 &   44.1 \\
\acrshort{yolo}v8-x & 36.3 & 3.56 &   68.7 \\
\bottomrule
\end{tabular}
\end{table}
\subsection{Post-Processing}
While \acrshort{od} models are the perfect choice for detecting objects in the scene, they don't provide any temporal information. Hence, to get that information and assign a tracking \acrfull{id} to each object, we have used state-of-the-art \acrfull{deepsort} \cite{wojke_simple_2017} model, which uses \acrfull{kf} \cite{chui_kalman_1987} to maintain tracking continuity from the previous state and predict the location of the bounding box in the subsequent frame. This algorithm leverages deep \acrshort{cnn} architecture in the prediction process, combining the strengths of \acrshort{kf} with the representational power of \acrshort{cnn}s. The object with the tracking \acrshort{id} can help distinguish between objects and provide the timestamps for the start and end of the object's life cycle.
\subsection{Results}\label{sec:results}
\begin{figure}
    \centering
    \includegraphics[width=\linewidth]{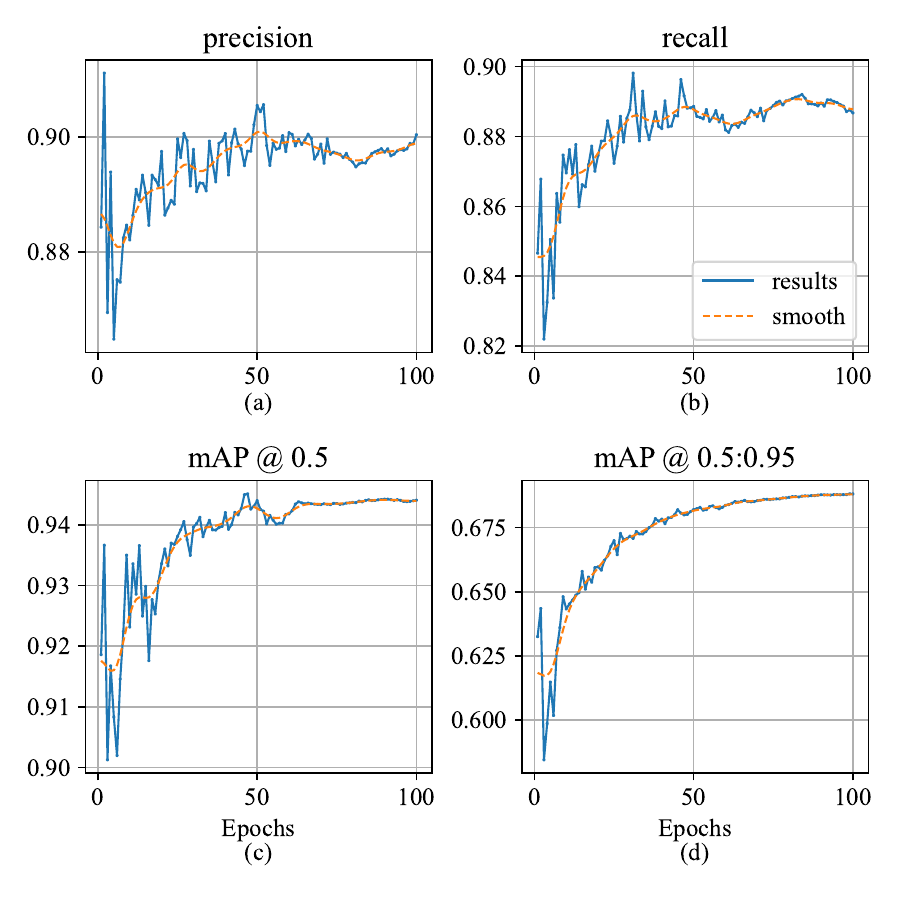}
    \vspace*{-5mm}\caption{Precision,  Recall, mAP@50 and mAP50-95 of the \acrshort{yolo}v8 \cite{jocher_yolo_2023} model training.}
    \label{fig:medium_results}
\end{figure}

We evaluated the performance of three variants (nano, medium, large) of the \acrshort{yolo}v8 model on detecting \textit{worker} and \textit{chair} classes in an industrial test set. The medium model exhibited superior overall performance with 94.4\% mAP@50, which is higher than both nano and large models with 1.8\% and 0.6\% respectively. Theoretically, the large model should have performed better due to more learnable parameters. However, as only two classes exist, more parameters can cause overfitting on the training set. Furthermore, a larger batch size helps better learning due to batch normalization as observed by \cite{you_large_2017, bjorck_understanding_2018}. 
We present the evaluation results of different trained model types in Table \ref{tab:model_performance}. In each column, items of the highest value are denoted in boldface. While the large model has a slightly better \acrshort{p} and \acrshort{r} on worker detection, it is compute-heavy. It takes 0.17 milliseconds (ms) more than the medium model on each frame, and it compounds up fairly when running in the real-time inference without adding significant value. Hence, we chose a medium model to conduct the analysis of the client's manufacturing facility and get more insights into the productivity of each station. 
Figure \ref{fig:medium_results} presents the training metrics of the medium model after each training epoch which represents mAP@50 is getting plateau after 65 epochs. \acrshort{p} and \acrshort{r} are always opposing metrics \cite{powers_2010_evaluation} and increasing \acrshort{p} has affected the \acrshort{r} as visible in the Figure \ref{fig:medium_results} (a,b) so that's why mAP provides a better measure to consider about the convergence of the model.
\begin{table}[htp]
\centering
\caption{Comparison of \acrshort{yolo}v8 \cite{jocher_yolo_2023} Nano, Medium and Large model performances. The highest values in each column are highlighted in boldface.}
\label{tab:model_performance}
\input{tables/results}
\end{table}
\section{Insights Into Manufacturing Facility} \label{sec:insights}
\begin{figure}
    \centering
    \includegraphics[width=1\linewidth]{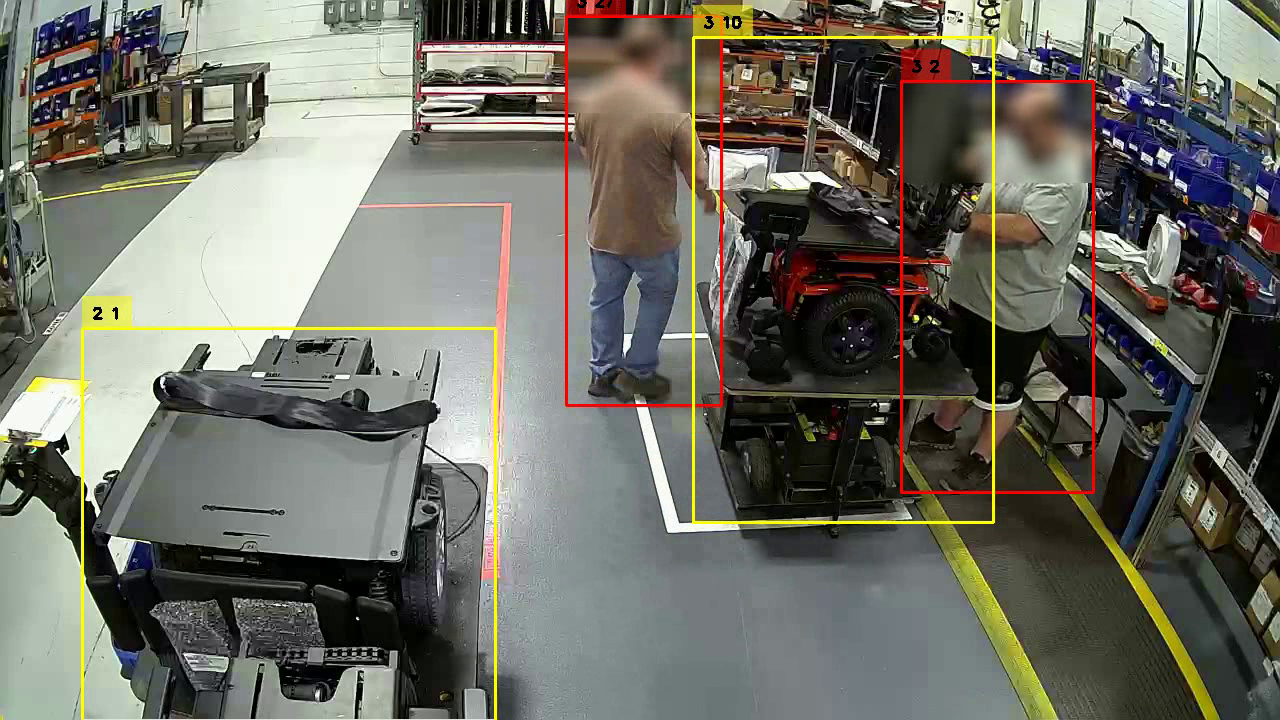}
    \vspace*{-5mm}\caption{The view of floor lines where power wheelchairs and workers are represented by yellow and red boxes, respectively.}
    \label{fig:view_line_A}
\end{figure}

The synergistic combination of \acrshort{yolo} and our proposed framework aims to highlight the specific challenges faced by the client in manufacturing assistive medical wheelchairs based on \acrshort{od}, localization, and tracking throughout the production process. 
With permission from the existing workers in the clients' facility, we strategically deployed cameras on the manufacturing floor to capture live video feeds for analysis. Figure \ref{fig:view_line_A} showcases the frame captured from an installed camera on-site with the view of a station where workers are assembling a chair.
\subsection{Challenges}
The wheelchair manufacturing process commences with specifications from healthcare professionals tailored to individual patient needs. These specifications dictate the assembly at various stations, each manned by a dedicated worker. Despite a standard processing time at each station, delays are common due to labor shortages and workers having to move between stations. Communication gaps exacerbate these delays, especially with floor managers who may be unaware of inventory issues. Transition periods, notably during shift changes, create inefficiencies and bottlenecks, impeding the production timeline. The presence of faulty items in the production line necessitates additional quality control time, disrupting the manufacturing flow. The inability to promptly identify and address bottlenecks, as indicated by data from the \acrfull{mes}, further delays production. The \acrshort{mes} data is often inaccurate, primarily reliant on manual worker input. Continuous manual time studies, while informative, are impractical and may lead to skewed productivity metrics.

\begin{figure}[!htbp]
    \centering
    \includegraphics[width=0.7\linewidth]{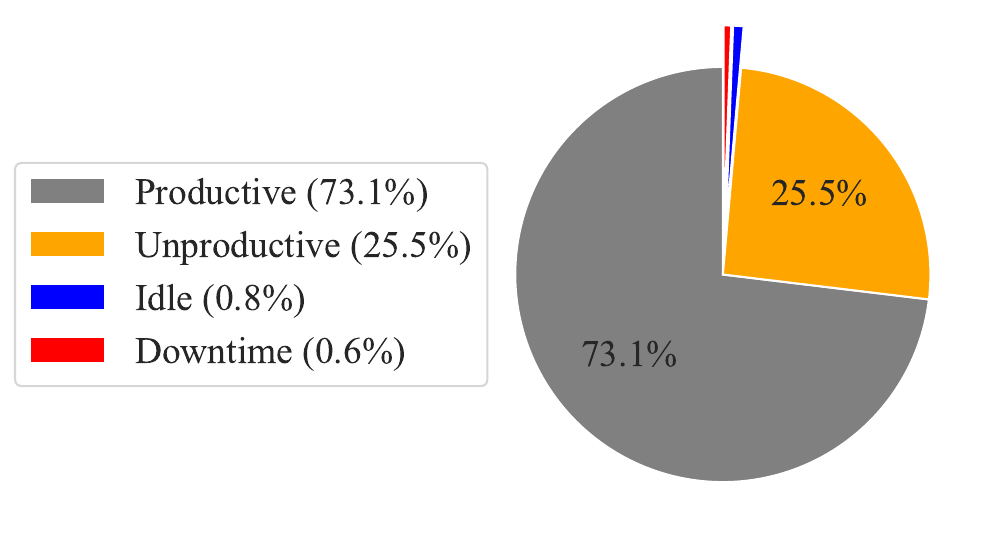}
    \vspace*{-5mm}\caption{The pie chart representing the status of Station C over 6 months.}
    \label{fig:productive_piechart}
    \vspace*{2mm}
    \centering \includegraphics[width=0.95\linewidth]{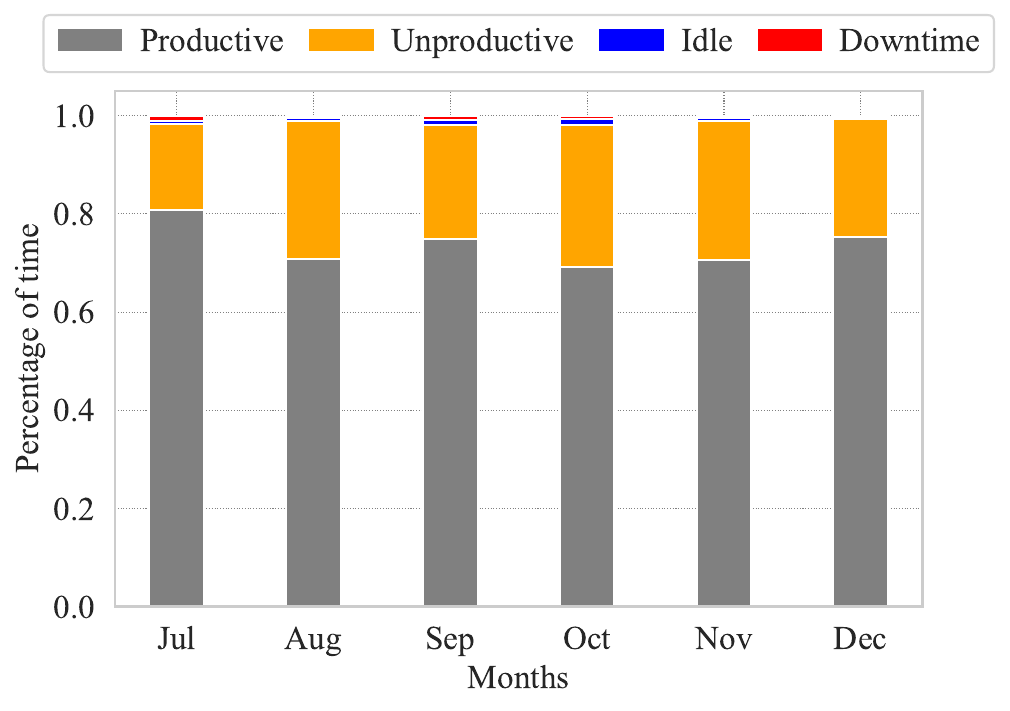}
    \vspace*{-5mm}\caption{Monthly aggregated status of Station C over 6 months.}
    \label{fig:monthly_percentage}
    \vspace*{2mm}
    \centering \includegraphics[width=0.95\linewidth]{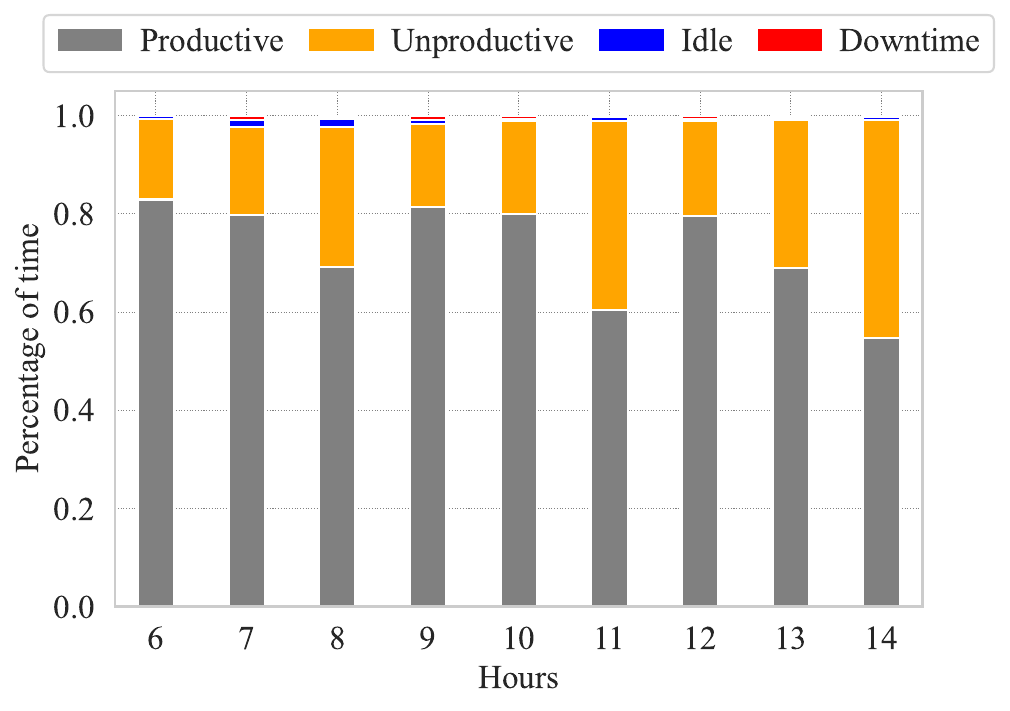}
    \vspace*{-5mm}\caption{Hourly aggregated data of the status of Station C over 6 months.}
    \label{fig:hourly_percentage}
\end{figure}

\begin{figure*}[!htbp]
    \centering \includegraphics[width=0.95\linewidth]{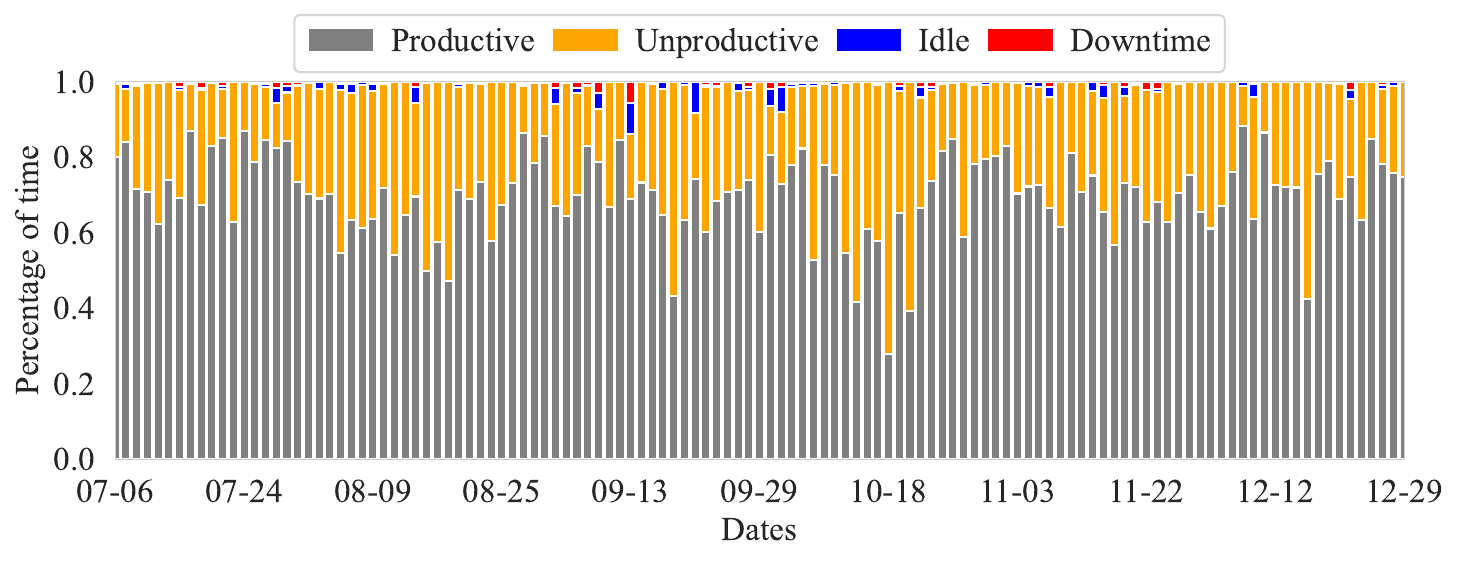}
    \vspace*{-5mm}\caption{Daily status of Station C in terms of productivity over 6 months.}
    \label{fig:daily_productivity}
    \centering \includegraphics[width=0.95\linewidth]{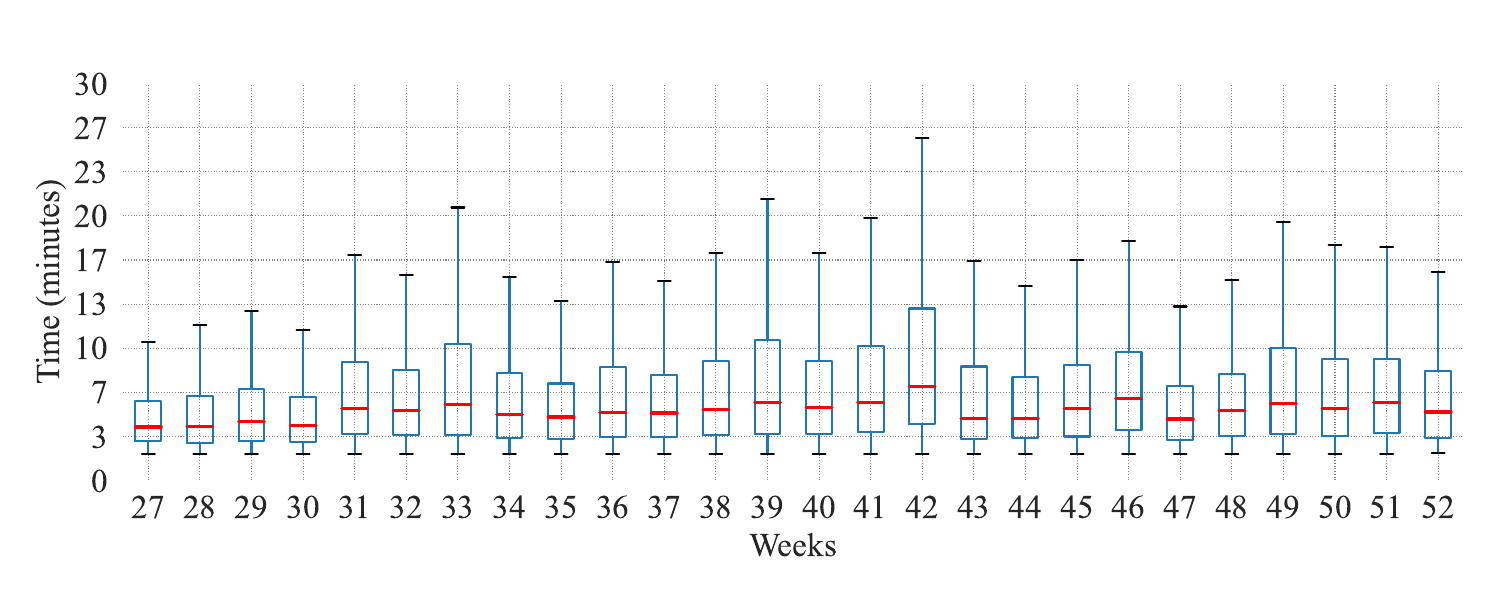}
    \vspace*{-5mm}\caption{Box plot of the processing time for all the chairs over 6 months aggregated for each week for station C. Red line, lower and upper box boundaries represents median and $25^{th}$ and $75^{th}$ percentiles respectively. }
    \label{fig:box_plot_grouped_by_week}
\end{figure*}

Given these challenges, we propose a non-invasive system for capacity constraint analysis. This system tracks the station and cycle time of chairs, allowing for effective labor and inventory planning. Over six months (July to December), we collected videos from four workstations (labeled Stations A, B, C, and D being the primary bottlenecks identified from \acrshort{mes} data while Station C being the critical one), excluding three standard 25-minute break times. The facility works only for the morning shift, so data was collected for 8.5 hours daily. We detected workers and chairs using the trained \acrshort{yolo}v8 model, as outlined in \S \ref{sec:labour_productivity}. Each frame's status was aggregated to extract various metrics. Figure \ref{fig:productive_piechart} shows that Station C is 27.9\% unproductive, indicating a critical labor shortage with a high percentage. Figure \ref{fig:hourly_percentage} depicts normalized hourly productivity data. Notably, productivity is higher in the morning than in the afternoon. Figure \ref{fig:monthly_percentage} reveals a consistent trend in productivity over the months, contrary to the expected increase towards the year's end due to high demand. This consistency points to a critical labor shortage. Figure \ref{fig:daily_productivity} represents the daily insights into their capacity constraint. They are productive only for 60\% to 65\% of time during the 8.5 hours shift and the remaining time is mostly unproductive when a worker is not available at the station to work on the chair. 

Further analysis of the life cycles of individual chairs is presented in Figure \ref{fig:box_plot_grouped_by_week}, where the box plot represents the five-number summary of the processing time of the chair each week. We have filtered out the processing time of less than two minutes due to the instances where a person is occluding the chair and the model is not able to detect it for a certain time. The tracker loses it and completes the life cycle and assigns a new id once the chair is visible assigning two distinct \acrshort{id}s for a same chair. It is evident that the median time (represented in red colored line) is reduced to 5 minutes on some days, while it also increases to 7.5 minutes (week 42), resulting in a 50\% increase in processing time and a decrease in daily output. Upon manual verification of the video data, the worker in charge was on vacation on week 42 and another worker from a different station was taking care of the station. A similar trend is also visible during the days when a worker handles multiple stations or during a severe inventory (week 39). 

These insights are crucial for workflow optimization. They are most effective when applied in a production environment with real-time metrics and an alert system for increasing processing time, as mentioned in \ref{fig:pipeline}. This system would enable floor managers to promptly address issues, thereby improving efficiency and productivity. While promising and groundbreaking results, the application of \acrshort{od} in this context also presents limitations in terms of tracking, especially when there is an occlusion or overlapping objects in the frame in the form of a worker occluding the chair being worked on for a longer period of time making it not visible or a guest visiting the station to casually chat for an extended period inflates the results. The safeguards around this will be explored in future studies.
\section{Conclusion} \label{sec:conslusion}
In this work, we implemented a non-invasive system for monitoring capacity constraints in manufacturing environments. Data was collected from four chair manufacturing assembly line stations over six months. After reviewing existing literature on \acrshort{od}, we propose a state-of-the-art end-to-end framework using \acrshort{yolo}v8 for object detection, providing insights into labor and inventory management, and revealing notable labor shortages and inefficiencies. The study underscores the importance of real-time metrics and alert systems in manufacturing environments to enhance efficiency and productivity. The overall productivity of Station C was calculated to be 70.6\% only over the course of 6 months suggesting a significant potential for technological integration in optimizing manufacturing processes. For future studies, we plan to use the \acrfull{lmm} to provide predictive, transcriptive, and prescriptive insights from the collected data. 
\section*{Acknowledgement}
A special thanks to Dario Morle and Syeda Sitara Wishal Fatima from IFIVEO CANADA INC.
for their helpful insights into the implemented methodology.
\bibliographystyle{IEEEtran}
\bibliography{IEEEabrv,references}

\begin{IEEEbiography}[{
\includegraphics[width=1in,height
=1.25in, clip, keepaspectratio]{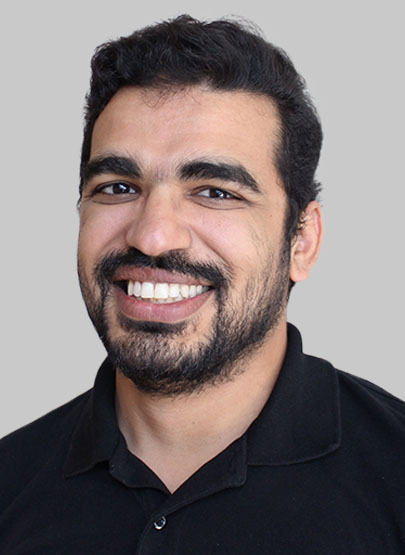}
}]
{Hafiz Mughees Ahmad} completed his Bachelor's and Master's in Electrical Engineering from the Institute of Space Technology, Pakistan, in 2015 and 2018, respectively. He is currently pursuing a Ph.D. at the University of Windsor, Canada. Alongside his studies, he serves as a Deep Learning Engineer at IFIVEO CANADA INC. His previous roles include Research Associate at Istanbul Medipol University, Turkey, and Lecturer at the Institute of Space Technology, Pakistan. His research focuses on Computer Vision and Deep Learning, with applications in OD and real-time surveillance and monitoring in the production environment. He is a Graduate Student Member of IEEE.
\end{IEEEbiography}
\vskip -2\baselineskip plus -1fil
\begin{IEEEbiography}
[{
\includegraphics[width=1in,height=1.25in,clip,keepaspectratio]{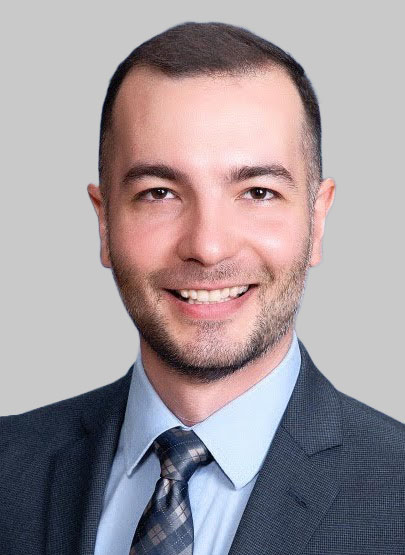}
}]
{Afshin Rahimi} received his B.Sc. degree from the K. N. Toosi University of Technology, Tehran, Iran, in 2010, and the M.Sc. and Ph.D. degrees from Toronto Metropolitan University, Toronto, ON, Canada, in 2012, and 2017, respectively, in Aerospace Engineering. He was with Pratt \& Whitney Canada from 2017 to 2018. Since 2018, he has been an Associate Professor in the Department of Mechanical, Automotive, and Materials Engineering at the University of Windsor, Windsor, ON, Canada. Since 2010, he has been involved in various industrial research, technology development, and systems engineering projects/contracts related to the control and diagnostics of satellites, UAVs, and commercial aircraft subsystems. In recent years, he has also been involved with industrial automation and using technologies to boost manual labor work in industrial settings. He is a senior member of IEEE, a lifetime member of AIAA, and a technical member of the PHM Society.
\end{IEEEbiography}
\vskip -2\baselineskip plus -1fil
\begin{IEEEbiography}
[{
\includegraphics[width=1in,height=1.25in,clip,keepaspectratio]{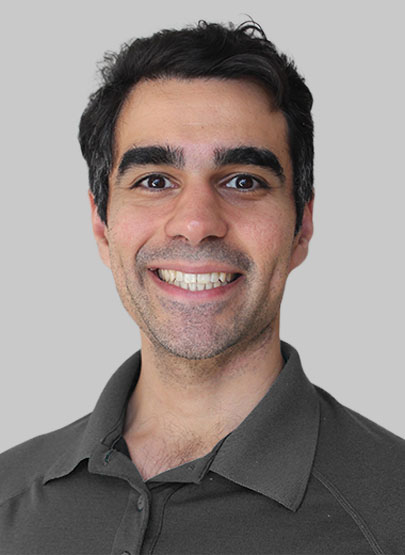}
}]
{Khizer Hayat} is the co-founder and Chief Technology Officer at i-5O. He is a Forbes 30 under 30 entrepreneur and has been working with enterprise startups for over 7 years. Currently, he leads product development at IFIVEO CANADA INC. His educational background is in AI and Robotics, having graduated with a Master's from the University of Pennsylvania. Previously, he spent 4+ years optimizing production processes using automation at Toyota Motor Corporation and Schlumberger Technology Corporation. He loves traveling and outdoor activities and is an avid Arsenal fan in his free time.
\end{IEEEbiography}
\end{document}

%% file: tables/results.tex
\begin{tabular}{lccccccc}
\toprule
\multirow{2}{*}{Model Size} & \multicolumn{3}{c}{P (\%)} &\multicolumn{3}{c}{R (\%)} \\ \cline{2-7}
 & All & Worker & Chair & All & Worker & Chair \\ \midrule
Nano   & 89.2 & 84.4 & 93.9 & 87.1 & 86.0 & \textbf{88.2} \\ \hline
Medium & \textbf{89.9} & 85.4 & \textbf{94.4} & 88.8 & 89.5 & 88.0 \\ \hline
Large  & 89.8 & \textbf{85.7} & 93.8 & \textbf{89.0} & \textbf{90.2} & 87.7 \\ \bottomrule

\end{tabular}
\begin{tabular}{lccccccc}
\toprule
\multirow{2}{*}{Model Size} & \multicolumn{3}{c}{mAP50 (\%)} & \multicolumn{3}{c}{mAP$_{50-95}$ (\%)} \\ \cline{2-7}
 & All & Worker & Chair & All & Worker & Chair \\ \midrule
Nano   & 92.6 & 91.7 & 93.5 & 64.7 & 64.8 & 64.6 \\ \hline
Medium & \textbf{94.4} &  \textbf{93.8} & \textbf{95.0} & 68.8 & 69.7 & \textbf{68.0} \\ \hline
Large  & 93.8 & 93.8 & 93.9 & \textbf{68.9} & \textbf{70.0} & 67.7 \\ \bottomrule

\end{tabular}